# The ND-IRIS-0405 Iris Image Dataset

Kevin W. Bowyer and Patrick J. Flynn

Department of Computer Science & Engineering
University of Notre Dame
Notre Dame, Indiana 46556

## I. Introduction

The Computer Vision Research Lab at the University of Notre Dame began collecting iris images in the spring semester of 2004. The initial data collections used an LG 2200 iris imaging system for image acquisition. Image datasets acquired in 2004-2005 at Notre Dame with this LG 2200 have been used in the ICE 2005 and ICE 2006 iris biometric evaluations [1,2]. The ICE 2005 iris image dataset has been distributed to over 100 research groups around the world.

The purpose of this document is to describe the content of the ND-IRIS-0405 iris image dataset. This dataset is a superset of the iris image datasets used in ICE 2005 and ICE 2006.

## II. Protocol for Image Acquisition

Images were acquired at the University of Notre Dame under an IRB-approved protocol. All subjects participating in the image acquisition signed a consent form at each acquisition.

### II.A. The LG 2200 System

All images in the ND 2004-2005 iris image dataset were acquired using the same LG 2200 iris biometrics system (see Figure 1). The LG 2200 uses near-infrared illumination of the eye, and provides audible prompts to help the subject position their head appropriately for image acquisition. The LG 2200 system illuminates the ocular region using three near-IR LEDS activated in sequence (one at a time) as it acquires images. These illuminators are positioned above, below-left, and below-right of the sensor. For most images, a specular highlight from the illuminator is clearly visible in the pupil region. The position of this highlight is a fairly reliable clue to which illuminator was active when the image was acquired.

The LG 2200 normally acquires a "shot" of three images with each illuminator active in exactly one of the images (see Figure 2). The standard camera software executes quality checks on these three images selects one of them for use, and discards the other two. These quality checks are performed by driver software. Through a cooperative research and development agreement with Iridian (which is now part of L1ID), software was supplied by Iridian to Notre Dame to allow all three images in a "shot" to be kept. Thus, one image our of each shot of three images acquired is guaranteed to pass the LG 2200's built-in quality control checks, and the other two may or may not pass the quality checks. This ability to keep images that do not meet the LG 2200's built-in quality control checks was intentional. The motivation was to have an image dataset that was not composed only of "pristine" images, but that could be used to support research on iris biometrics algorithms that could cope with non-ideal image quality.

## I.B. Internal Contrast Enhancement

The LG 2200 system acquires images by digitizing an NTSC video signal, returning an image with 480 rows and 640 columns. The Iridian software used in our image acquisition saves a contrast-stretched version of the digitized image. Each image has some pixels that have the intensity value 0 and some pixels that have the intensity value 255. In addition, every third intensity value is unused. Figure 3 depicts a representative histogram of gray values from an iris image as stored by the LG2200 system.

## I.C. Motion-induced interlace artifact

The LG 2200 is an active image acquisition system in several senses. One, the illumination source continually rotates between three illuminators. Two, the camera continually modifies its focus setting to maximize texture detail in the iris. Three, the human subject typically moves to position their eye correctly for image acquisition. The fact that the system is designed to incorporate elements of motion, combined with the fact that the image is digitized from an NTSC signal, means that scan-line interlace artifacts will be visible in some images. These interlace artifacts are a consequence of interlaced (field-based) sensing (see Figure 4).

## II. Description of Subject Population

The ND 2004-2005 iris image dataset contains 64,980 images corresponding to 356 unique subjects, and 712 unique irises. The age range of the subjects is 18 to 75 years old. 158 of the subjects are female, and 198 are male. 250 of the subjects are Caucasian, 82 are Asian, and 24 are other ethnicities. None of the images correspond to subjects wearing glasses during image acquisition. However, a significant fraction of the subjects wore contact lenses. Image artifacts arising from contact lenses are discussed in a later section.

## III. Welcome to Real-World Images

The ND 2004-2005 iris image data acquisition is large enough that many real-world conditions occur that may present challenges for iris biometrics systems. There are some images that have noticeable blur, and some in which the iris region is substantially occluded; see Figure 5 for examples of these conditions. There are some images in which the iris region is cut off at the image boundary; see Figure 6 for examples of this. While most iris images are acquired in a near straight-on view, another real-world condition that occurs is 'off-axis' images, where the subject's view direction is not anti-parallel to the optical axis of the camera. This happens even though the LG 2200 is designed to encourage collection of straight-on images. See Figure 7 for examples of off-axis views of the iris. Contact lenses can present a wide variety of artifacts in an image. Cosmetic lenses are an obvious challenge. Hard lenses generally present an obvious refractive boundary. Soft lenses can be difficult to spot when present, but often the rim of the lens may be detected as a contour on the sclera outside of the limbus. See Figure 8 for examples of artifacts resulting from contact lenses. Additional discussion and examples of contact lens artifacts appears in [3,4].

**IV. Publications Using Subsets of ND-IRIS-0405 Data**

Subsets of the ND-IRIS-0405 iris image dataset have been used in a large and growing number of research publications [5-21]. The ICE 2005 subset is the most widely used to date.

**V. Relation of ND-IRIS-0405 Dataset to ICE 2005 and ICE 2006 Datasets**

The ND-IRIS-0405 dataset is a superset of ICE 2005 and ICE 2006. That is, there are images in ND-IRIS-0405 that are not in either of the ICE 2005 or ICE 2006 datasets. At the time that the ICE 2005 dataset was released, the ICE program guidelines called for the filenames to be "anonymized". This was done so that iris images could not be cross-linked to face images used in the FRGC or FRVT programs. Image filenames in ND-IRIS-0405 are not anonymized in this way. The data archive contains a file that gives the correspondence between ICE 2005 filenames and ND-IRIS-0405 filenames. The subset of ND-IRIS-0405 that corresponds to the ICE 2006 dataset is currently not identified, pending its use in other government programs.

**VI. Updates to the ND-IRIS-0405 Dataset**

Our efforts to eliminate errors with identity labeling and metadata in the database are extensive. Our error detection procedure employs both human inspection and processing with commercial iris recognition software. However, it is possible that there are labeling errors (e.g., wrong subject ID in the file name) or metadata errors (e.g., wrong eye listed) in the data set. We are committed to fixing all errors. We are also committed to incorporating all images collected between January 2004 and May 2005 in this data set, including those withheld from this initial release because of labeling problems. In keeping with this commitment, this data set may be updated. The data set will have a version number and a CHANGELOG.txt file that describes the updates made between versions. Licensees will be informed of updates to the data set and given the opportunity to pick up the changes. This process is made easier if licensees keep a pristine copy of the downloaded data at their site, from which working copies are made and used for the licensee's R&D efforts. Changes made to the database are then obtained by running another rsync download with appropriate flags ("--delete", for UNIX and Mac clients) to remove files locally that have been renamed or deleted at the server.

**Acknowledgements**

Biometrics research at the University of Notre Dame has been supported by National Science Foundation grant CNS01-30839, by the Central Intelligence Agency, by the Intelligence Advanced Research Projects Activity and by the Technical Support Working Group under US Army contract W91CRB-08-C-0093. The opinions, findings, and conclusions or recommendations expressed in this publication are those of the authors and do not necessarily reflect the views of our sponsors.

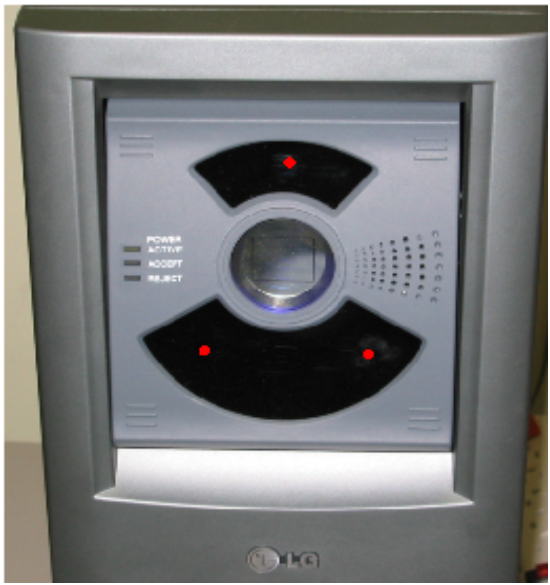

**Figure 1 – The LG 2200 Iris Imaging System. Red dots have been added to indicate the LED illuminant positions.**

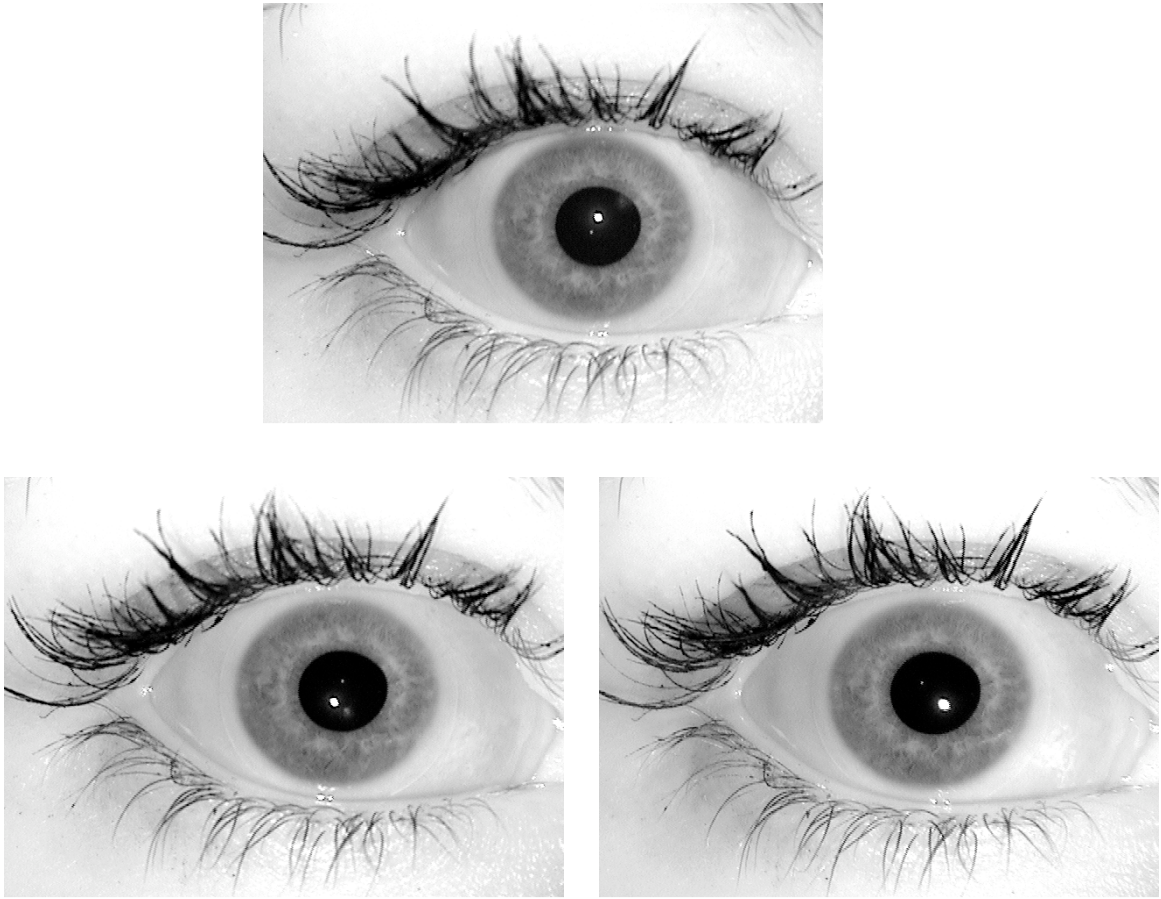

**Figure 2 – An Example "Shot" of Images Acquired Using the LG 2200.**

Images 04213d382 (top), 04213d383 (left) and 04213d384 (right) represent one "shot" from the LG 2200. The illumination for each image is provided by a different one of the three near-IR LEDs. The relative position of the LEDs on the system is apparent here in the relative position of the specular highlights in the pupil region of the images. These images are a part of the ICE 2005 dataset.

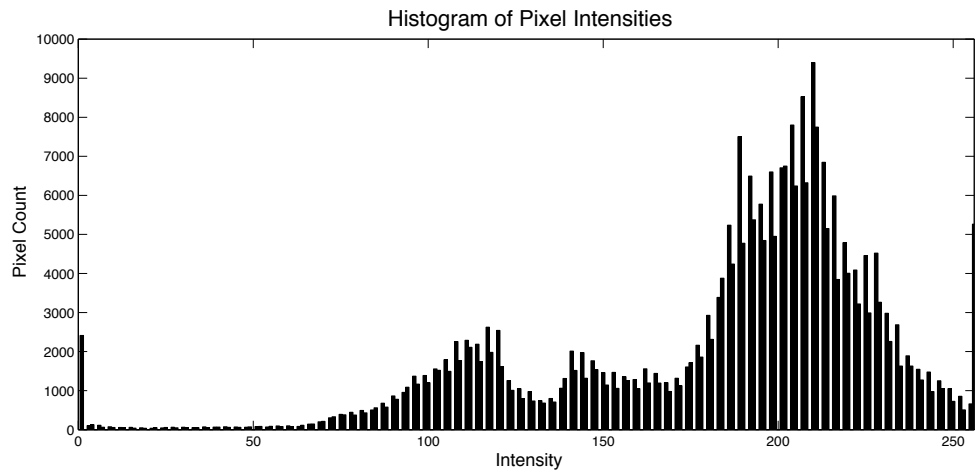

**Figure 3 – Effect of Internal Contrast Enhancement on Image Histogram.**

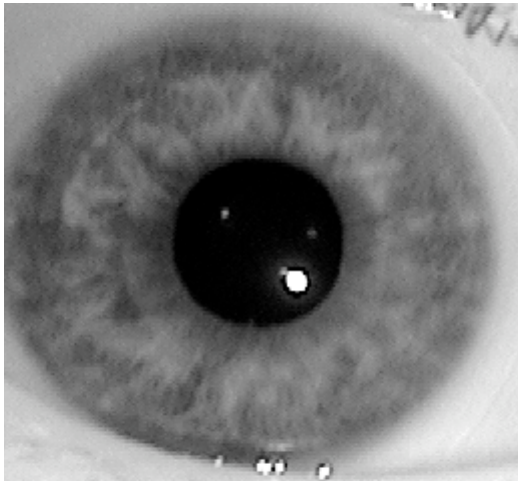 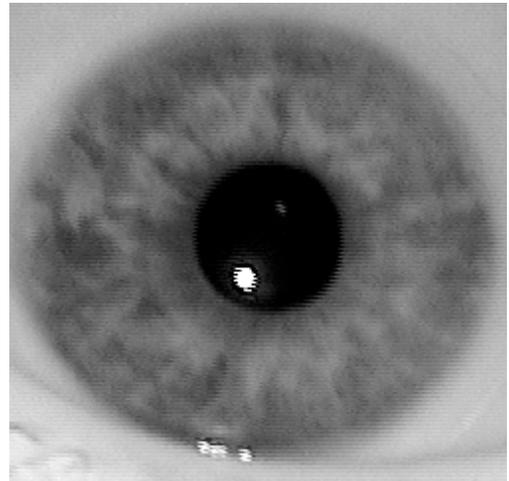

(a) (b)

**Figure 4 – Example of Motion-Induced Interlace Artifact.
(a): Iris area of image 04261d455.tiff. (b): Iris area of image 04261d456.tiff, showing interlace offset due to motion.**

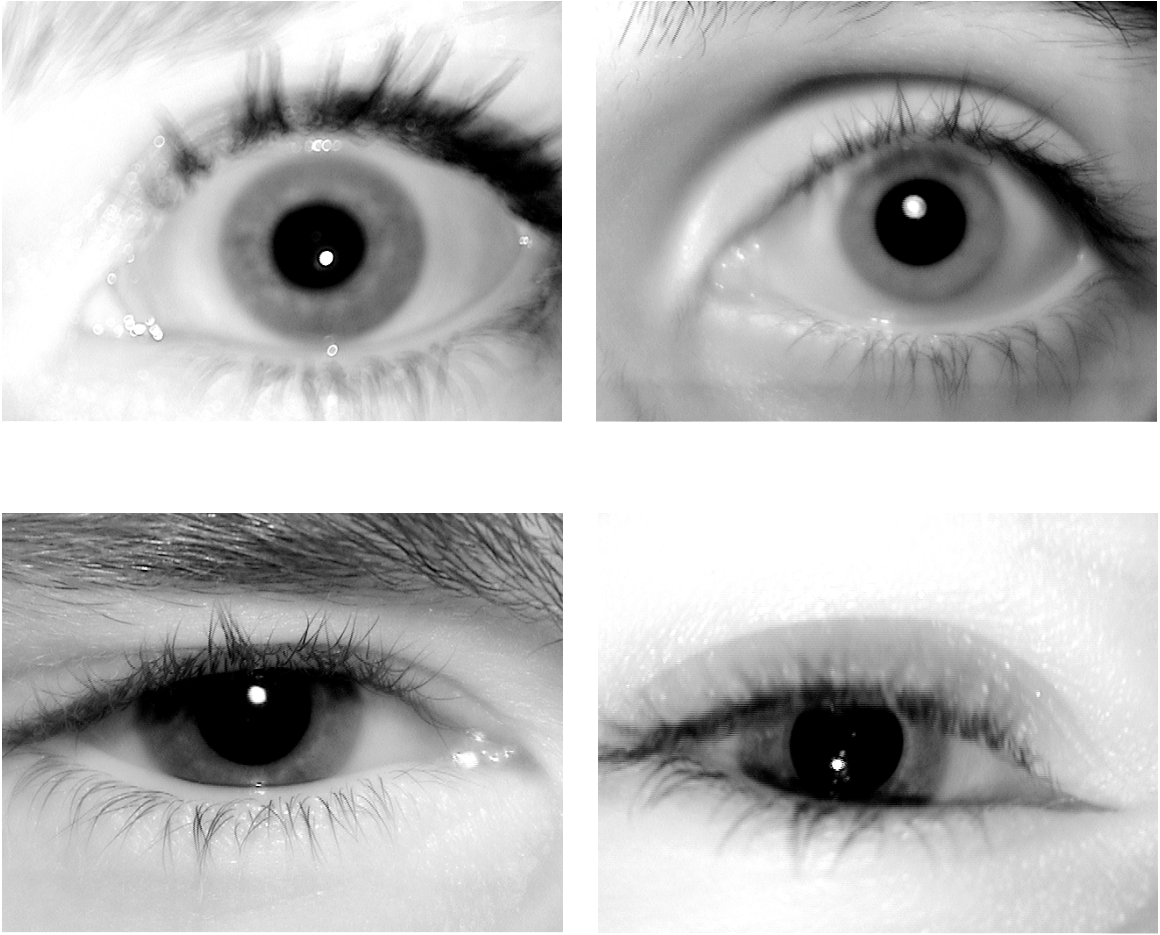

**Figure 5 – Examples of Out-of-Focus and Occluded-Iris Images.**

Images 04213d356 (upper left) and 04395d278 (upper right) exhibit significant blur. Images 04915d106 (lower left) and 04787d76 (lower right) exhibit significant occlusion of the iris region by some combination of eyelids, eyelashes and shadows. These images are a part of the ICE 2005 dataset.

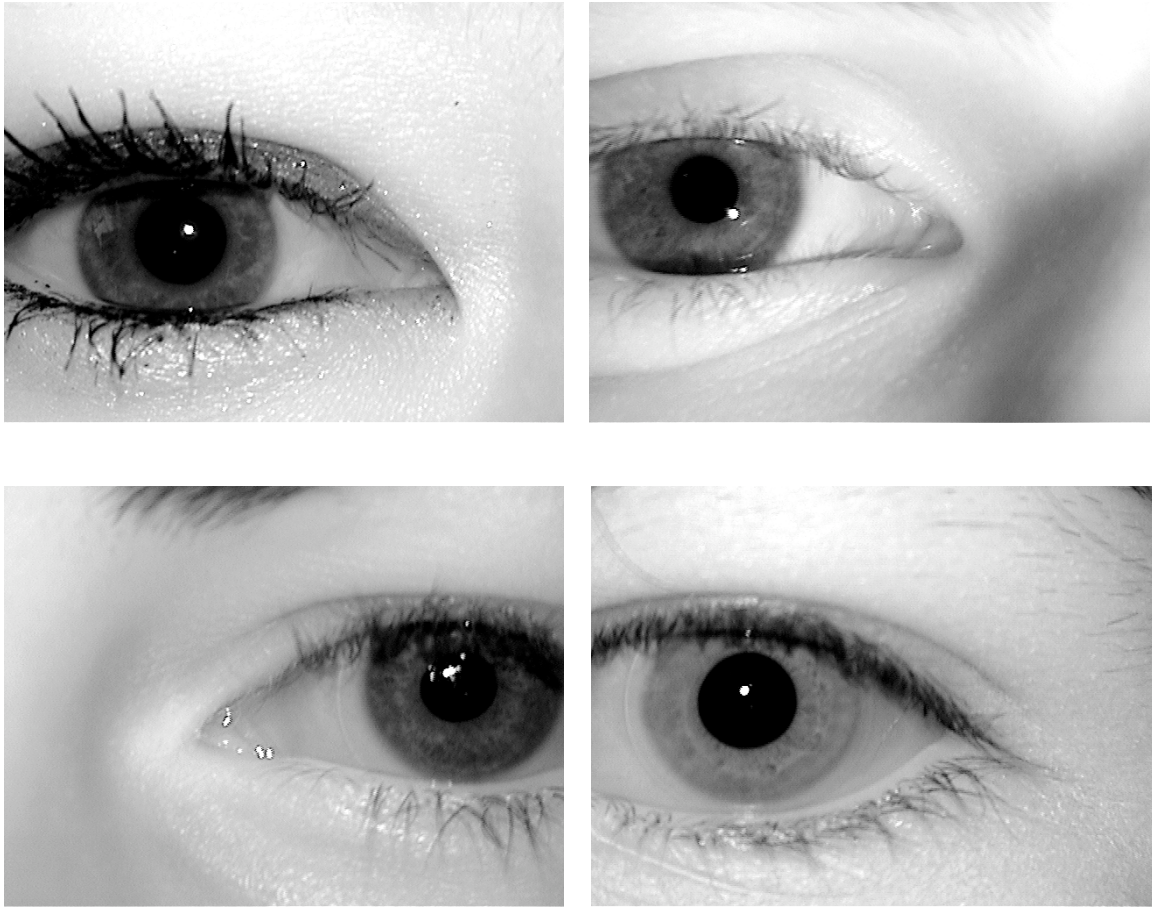

**Figure 6 – Example Images With Eye Region Cut Off by Image Boundary.**

Images 04920d09 (upper left), 04932d85 (upper right), 04888d55 (lower left) and 04767d126 (lower right) have some fraction of the eye region that does not appear in the image. These images are a part of the ICE 2005 dataset.

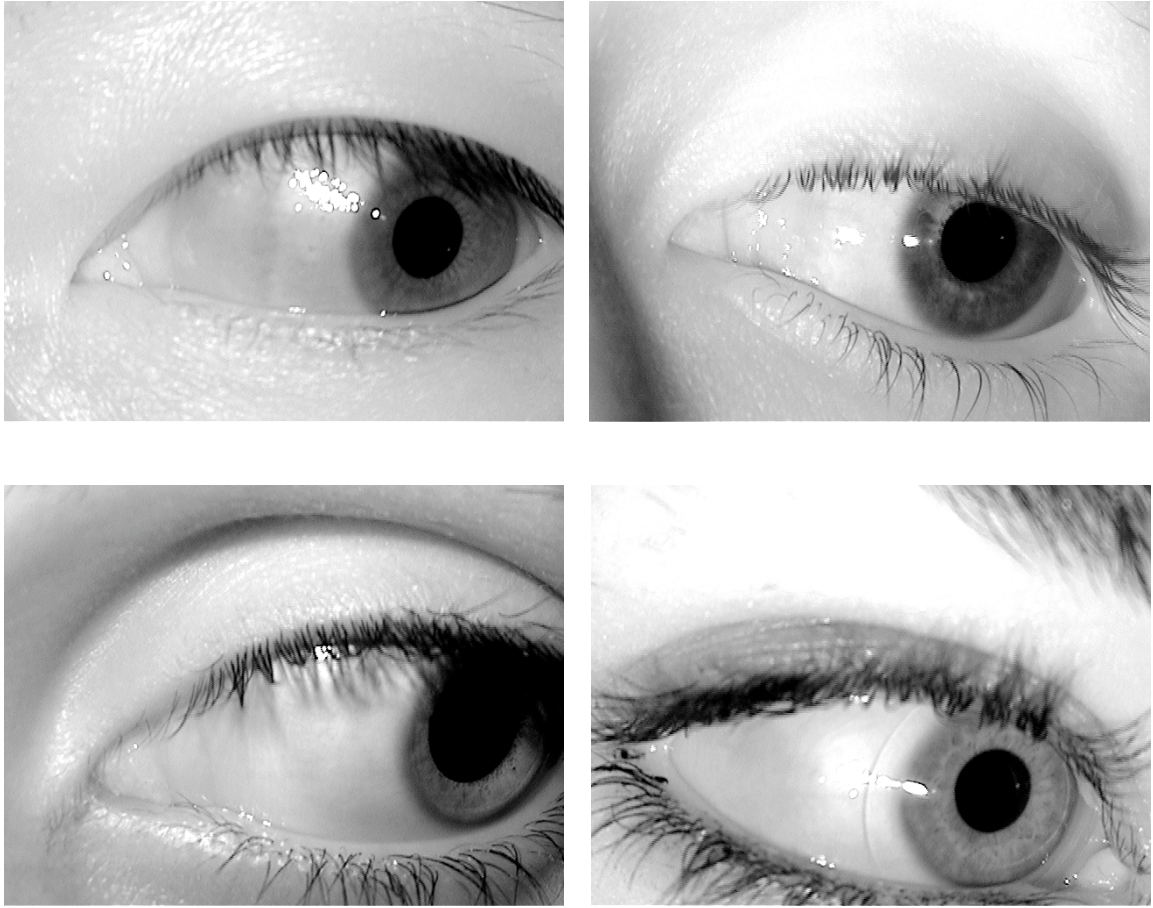

**Figure 7 – Example Images With Off-Axis View of the Iris.**

The view of the iris region is substantially off-axis in images 04936d117 (upper left), 04609d224 (upper right), 04395d281 (lower left) and 04863d77 (lower right). These images are a part of the ICE 2005 dataset.

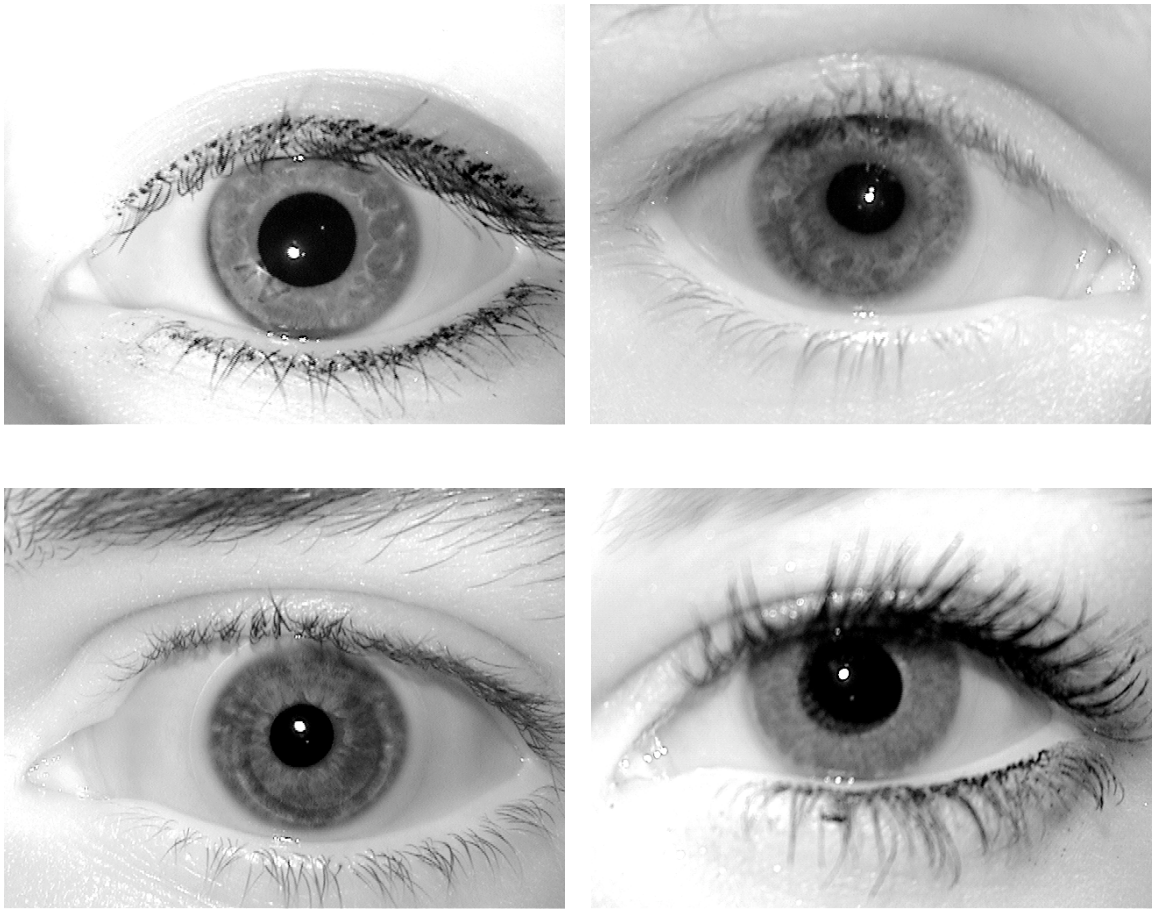

**Figure 8 – Example Images With Artifacts Resulting from Contact Lenses.**

Image 04887d100 (upper left) has the "AV" lettering visible in the lower left of the iris. Image 04869d104 (upper right) has a hard contact lens shifted over the lower part of the iris region. Image 04456d375 (lower left) has concentric banding slightly off-center of the pupil. Image 04780d115 (lower right) shows a cosmetic contact lens. These images are a part of the ICE 2005 dataset.